\documentclass{article}

\usepackage{arxiv}
\usepackage{caption}
\usepackage{graphicx}
\usepackage[utf8]{inputenc} 
\usepackage[T1]{fontenc}    
\usepackage{hyperref}       
\usepackage{url}            
\usepackage{booktabs}       
\usepackage{amsfonts}       
\usepackage{nicefrac}       
\usepackage{microtype}      
\usepackage{lipsum}
\usepackage{listings}
\usepackage{courier} 
\usepackage{listings, xcolor}
\usepackage{caption} 
\title{Constructing a personalized learning path using genetic algorithms approach}

\author{
  Lumbardh Elshani, Krenare Pireva Nuçi  \\ 
  Department of Computer Science and Engineering\\
  University for Business and Technology\\
  10000 Prishine, Kosovo \\
  \texttt{le41207@ubt-uni.net, krenare.pireva@ubt-uni.net} \\

}

\begin{document}
\maketitle

\begin{abstract}
A substantial disadvantage of traditional learning is that all students follow the same learning sequence, but not all of them have the same background of knowledge, the same preferences, the same learning goals, and the same needs. Traditional teaching resources, such as textbooks, in most cases pursue students to follow fixed sequences during the learning process, thus impairing their performance. Learning sequencing is an important research issue as part of the learning process because no fixed learning paths will be appropriate for all learners. For this reason, many research papers are focused on the development of mechanisms to offer personalization on learning paths, considering the learner needs, interests, behaviors, and abilities. In most cases, these researchers are totally focused on the student's preferences, ignoring the level of difficulty and the relation degree that exists between various concepts in a course. 
This research paper presents the possibility of constructing personalised learning paths using genetic algorithm-based model, encountering the level of difficulty and relation degree of the constituent concepts of a course. 
The experimental results shows that the genetic algorithm is suitable to generate optimal learning paths based on learning object difficulty level, duration, rating, and relation degree between each learning object as elementary parts of the sequence of the learning path. From these results compared to the quality of the traditional learning path, we observed that even the quality of the weakest learning path generated by our GA approach is in a favor compared to quality of the traditional learning path, with a difference of 3.59\%, while the highest solution generated in the end resulted 8.34\% in favor of our proposal compared to the traditional learning paths.
\end{abstract}


\section{Introduction}
\label{sec:Intro}
The rapid changes of technology and the increase in the internet usage has affected many life processes, and one of them that has been greatly affected is the learning process, as one of the most sensitive processes in the cycle of human activity. Learning is the act of acquiring new knowledge, skills, behaviors, or values, or modifying and reinforcing them, and can involve the synthesis of different types of information. This ability is possessed by humans, animals, and some types of machines. Learning is a continuous process that occurs throughout a person's life, but the trajectory followed by everyone is not suitable for everyone equally. Hence, the need to personalize learning paths is essential to increase performance and effectiveness of the knowledge gained for a learner. But first, to understand the importance of providing personalization in the learning paths, we must understand what a learning path is? A learning path can be described as a chosen path, taken by a learner through a variety of learning activities, allowing him / her to build knowledge progressively \cite{pireva2017representation}. 

Studies on learning paths help us to explore and explain student behaviors during learning processes, since they have unique knowledge structures based upon their previous experiences and abilities. Learning paths also reveal the learning trails while learners traverse any interactive environment \cite{jih1996impact}. A learning path allows people to gain knowledge in small chunks, giving students the role of helmsman. It can be customized in the way it creates self-taught and self-paced learning experiences. Now, we need to familiarize ourselves with personalized learning. Personalized learning refers to a wide variety of educational programs, academic strategies, learning experiences, and systematic pedagogical approaches designed to address student’s specific learning needs, aspirations, and interests on an individual basis. Whereas the general purpose is to put in the primary consideration the individual learning needs. The United States National Education Technology Plan 2017 defines personalized learning as follows\cite{thomas2016future}:
“Personalized learning refers to instruction in which the pace of learning and the instructional approach are optimized for the needs of each learner. Learning objectives, instructional approaches, and instructional content (and its sequencing) may all vary based on learner needs. In addition, learning activities are meaningful and relevant to learners, driven by their interests, and often self-initiated”.
The purpose of personalized learning is to facilitate academic success for each student by first defining the needs, interests, and aspirations of students, then continuing to provide personalized learning experiences for each of them. The aim of this paper is to provide a genetic-based model to generate personalized learning paths considering course difficulty level and the relation degree that exists between different concepts of that course. Also, the approach presented in this paper tends to use different population initialization methods and different genetic operators, so that the final results could be comparable in the end. The paper is structured as follows: In section 2, is presented the related work for personalized learning paths. Followed by methodology and experimental design. Before the conclusion, the results and discussion are presented. 

\section{Related Work}
\label{sec:lr}
In this section are presented the use of several techniques to provide personalised learning, followed by the specific technique, such as the use of genetic algorithms.

\begin{figure}[h]
    \begin{center}
        \includegraphics[width=150mm]{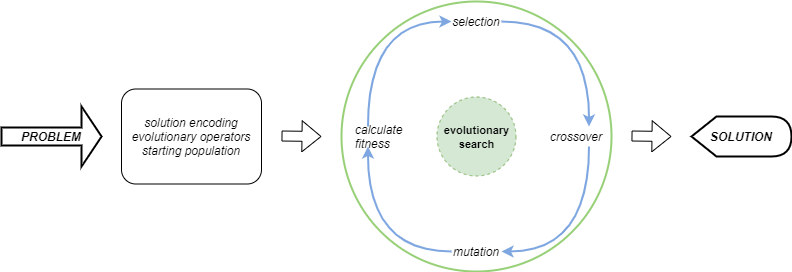}
        \caption{The basic cycle of Evolutionary Algorithms}
        \label{fig:evolutionary-algorithms}
    \end{center}
\end{figure}

Lately researches, have experiment with several techniques to generate adaptive learning paths. Such techniques as Learning Path Graph\cite{karampiperis2005adaptive}, Multi Agent Systems\cite{pireva2014cloud}, Artificial Intelligent Automated Planning\cite{pireva2017recommender}, Ontology \cite{gascuena2006domain}, Swarm Intelligence\cite{de2007competency}, Neural Networks\cite{kwasnicka2008learning}, Bayesian Networks \cite{kumar2005rating, neil2005using, heckerman2008tutorial}, Evolutionary Algorithms (EA) \cite{eiben2003introduction}, to name a few. The latter one is a group of algorithms based on the evolution theory of Charles Darwin \cite{eiben2003introduction}. The principle followed by these algorithms is known as survival of the fittest, which implies that environmental pressure causes natural selection in a population of individuals. As a class of general, random search heuristics, these can be applied to many different tasks. EAs tend to find optimal solutions within specific constraints by mimicking biological mechanisms, such as mutation, recombination, and natural selection.
Figure \ref{fig:evolutionary-algorithms} describes how all evolutionary algorithms work. First, a real-life problem must be represented in a machine-readable form, then the algorithm starts working with a population whose individuals are randomly generated. From this population, each candidate solution is evaluated and selected according to its fitness and then their recombination is done to form optimal solutions.

\subsection{Personalized Learning Path using Genetic Algorithm approach}
\label{sec:plpga}
Genetic algorithm (hereafter: GA) is usually used as an optimization technique to search the global optimum of a function and due to the huge search space, the optimization problem is often very difficult to solve. GAs have shown a good performance a wide variety of problems. In \cite{de2007competency} for personalising learning paths GA is modelled as a classical constraint satisfaction problem, and since then, research on usage of GAs to offer personalized learning has emerged. Jebari et al. (2011), in his research paper presented a solution based on GAs to find suitable dynamic personalized learning sequences of exercises, respecting the constraints, and maximizing success of the student \cite{jebari2011genetic}. 

To determine an optimal learning path based on GAs, different approaches have been explored with respect to personalised learning paths. As analysed from the paper \cite{huang2007constructing}, GA is applied to generate personalized learning paths based on case-based reasoning, considering curriculum difficulty level and continuity of successive curriculums. Here, authors have used a database to store course information along with the coefficient of difficulty. Then, based on the test results, the appropriate courses are selected, so that they require the lowest level of work coefficient. Finally, a GA generates an optimal individual training program for each student using the data obtained in the selection of courses ranked according to the work coefficient. 
Moreover in \cite{chen2008intelligent}, authors applied GAs to propose personalized curriculum sequencing based on pre-test and pre-learning performance. The pre-test is based on incorrect test responses of each individual learner, which is performed at the start of the lesson to collect incorrect course concepts of learners through computerized adaptive testing, then the personalized mechanisms rely on the results of this pre-test. Furthermore, Bhaskar et al. (2010), used GAs to evolve learning path generation into a learning scheme generation. This learning scheme accommodates each student’s entire context, including their profile, preferences and learning contexts. Depending on the learning goals and intentions of the student, the right content is selected, which could be of different types: presentation, media, content. So, this approach has considered the content of learning. The GA in this approach is used to select the type of content that best suits the needs and goals of a student \cite{bhaskar2010genetic}. In Spain, a team of researchers from the University of Alcala has conducted research on the dynamic selection of learning objects using GA to construct a learning path depending on the learner’s competencies as an input and the planned learning outcomes as an output \cite{clement2000model}. The problem with the approaches mentioned above, is that they do not generate personalized learning paths adapted to each individual student, but just orient the student to navigate through a range of learning activities \cite{de2011genetic}. In \cite{hong2005personalized}, a group of researchers from an education university in Taiwan, have worked on constructing a personalized e-Learning system based on case-based reasoning approach and GA. Unlike most other works that consider only the interests, preferences and behaviors of students, the authors in their work \cite{hong2005personalized}, have considered the level of difficulty of the learning units and the ability of the student. Thus, their approach is based on the technique of evolution through computerized adaptive testing (CAT), then GA and case-based reasoning (CBR) are used to build an optimal learning path for each student. In \cite{zaporozhko2018genetic}, the authors aim to construct an optimal individual educational trajectory, based on GA, being as close as possible to the features and real possibilities of each listener of the online course, with the possibility of real-time correction. In addition, they continue to formulate the problem, built into the structure of a massive open online course (MOOC), trying to provide solutions by classifying students into 4 groups, differing in the dominant learning style: visual learners, aural learners, read-write learners and kinesthetic learners and formulating a mathematical model. On this model the generation of individual learning route continues, applying the sequence of genetic algorithm steps to optimize the learning route. At the end, the obtained results are evaluated, and the generated solutions are analyzed against the optimal solutions. Moreover, Bellafkig et al. (2010), conceives an adaptive educational system based on the modeling of the description of pedagogical resources. In this paper, this raised problem is considered as a problem of optimization, and to build a sustainable solution the proposed system uses genetic algorithms to seek for an optimal path that starts from the student profile towards pedagogical objectives, passing through intermediate courses. The contribution of this paper is the construction of a system architecture, the adaptation of algorithms in this system and the modeling of a standard content module. The author through this paper targets important parts for the complete construction of an adaptive learning system, starting with the definition of the system architecture, which is generally divided into two parts: the student space where his identifiers fit, and the space of expert seeking the integration of new resources into the system. The work then continues with the adaptation module, which includes GA and adapter functionality, to proceed then with the implementation of the scenario in modeling and adaptation. The general structure of the architecture and adapters is presented with the help of diagrams, while the modeling part of the learning object is realized through the multi-part standard, to facilitate research, evaluation, reuse, and acquisition, for example by students, instructors, or automated software. To meet the conditions of the presented approach, the author assumes that the student has a precise intellectual baggage, being able to understand a certain course and this set of knowledge is called as pre-requisite concepts. On the other hand, each course is accompanied by a certain set of knowledge to be gained after the course, and this is called as post-acquired concepts \cite{bellafkih2010adaptive}. 

However, from the aforementioned research works, our work differs in the proposed approach of building a hybrid system to generate optimal learning paths considering factors such as: topic difficulty level, duration, rating, and relation degrees of topics, and it aims to contribute on increasing the quality of learning path generated, as well as increasing the performance of the system by developing different techniques for population initialization, selection and recombination, which will be discussed further in the following sections.

\section{Methodology and Design of Genetic Algorithm Model}
\label{sec:mde}

The general steps to follow when applying GA in optimizing a particular problem are: (i) Population Initialization, (ii) Fitness Calculation, (iii) Selection, (iv) Crossover, and (v)  Mutation \cite{holland1992adaptation}.
Being part of EA, GA before starting the process of searching for optimal solutions from the search space, must have defined two things, which are also the basic requirements of a GA:

\begin{enumerate}

\item \textbf{Solution Encoding} - Indicates the representation of a solution within the computation space. This is a process of transforming a solution from the phenotype (representation of actual real-world solution) to genotype space (representation of a solution in computation space that can be easily understood and manipulated using a computing system) \cite{rothlauf2009representations}. This process is illustrated in Figure \ref{fig:solution-encoding}.

\begin{figure}[h]
    \begin{center}
        \includegraphics[width=70mm]{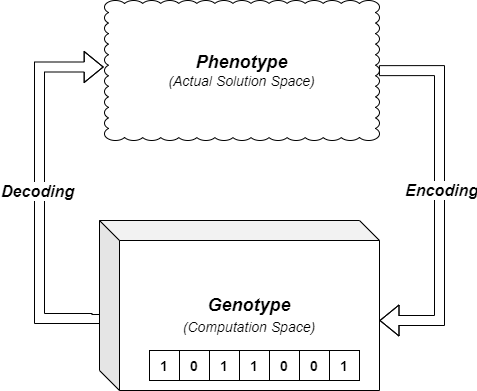}
        \caption{The process of encoding and decoding a solution}
        \label{fig:solution-encoding}
    \end{center}
\end{figure}

To model an individual solution to the system, we used integer-based encoding because this way of encoding is easier to apply then recombination and selection operators. An integer number is assigned to each concept from 0 to n - 1, where n is the maximum number of courseware concepts.
Figure \ref{fig:our-solution-encoding} contains a visual description of the solution modeling, where in each cell is the ID number that determines each concept in a courseware.

\begin{figure}[h]
    \begin{center}
        \includegraphics[width=120mm]{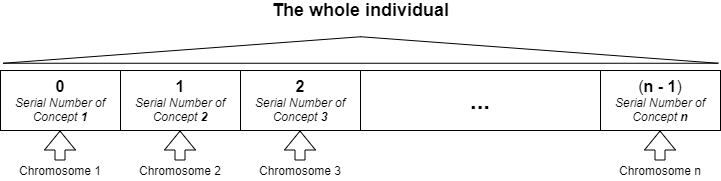}
        \caption{Solution Encoding}
        \label{fig:our-solution-encoding}
    \end{center}
\end{figure}

The structure of the solution presented on Figure \ref{fig:our-solution-encoding} determines a sequence of interrelated concepts, and as such is evaluated by an objective function, to judge its quality based on the level of difficulty of each concept and the relation degree with the next concept.

\item \textbf{Fitness Function} - Determines which solutions within a population are better at solving the problem under consideration. This function simply takes a solution as input and indicates how suitable that solution is regarding the studied problem, and it should be highly optimized and must evaluate the quality of a solution in quantitative values.

To generate personalized learning path for individual learners based on their pretest results, difficulty level, courseware concepts relation degrees, rating and concept granularity are considered in our method. To evaluate how fit and close a generated solution is to the optimum solution of personalized learning path generation problem, we mathematically modeled the fitness function as follows:

\[ \textbf{\textit{f}} =  \sum_{i=1}^{n} ((\frac{g_{i}}{r_{i}}) * (1 - w) * (rd_{(i-1)i})) + w * (1 - d_{i})  \]

where \textbf{\textit{f}} is the fitness function used by GA to evaluate each solution, \textbf{g}\textsuperscript{i} is the concept duration or granularity, \textbf{r}\textsuperscript{i} is the concept rating, telling how useful was the particular concept for that particular learner, \textbf{rd}\textsubscript{(i-1)i} represents the \textbf{(i - 1)}\textsuperscript{th}  courseware concept relation degree with the \textbf{i}\textsuperscript{th} concept in the learning path, \textbf{d}\textsubscript{i} is the difficulty parameter of the \textbf{i}\textsuperscript{th} concept, n stands for the total number of concepts that build a courseware, while w is an adjustable weight. The code written to model the fitness function formulated above is presented in Listing \ref{lst:fitness-code}. RDM stands for Relation Degree Matrix, which is a two-dimensional array containing relation degrees of each courseware concept with others.

\begin{lstlisting}[label={lst:fitness-code}, caption = {Fitness Function}, captionpos=b, language = Java , frame = single, escapeinside={(*@}{@*)}]
private double calculateFitness() {
    double fitness = 0;
    for (int i = 1; i < solution.size(); i++) {
        Concept c = course.getConcept(solution.get(i));
        Concept pC = course.getConcept(solution.get(i - 1));
        fitness += (c.getGranularity() / c.getRating()) * 
		(w * RDM[c.getID()][pC.getID()]) + ((1 - w) * c.getDifficulty());
    }
    return (double)Math.round(fitness * 100) / 100;
}
\end{lstlisting}

\end{enumerate}

Immediately after the solution representation and fitness function are defined, GA begins to initialize a population of randomly solutions, or using other combined approaches. Then the population improvement is done by repeatedly applying recombination operators and selection techniques. After each iteration, a new generation of population with individuals is created and in each generation the fitness of each solution is evaluated. The process of creating new generations and evaluating solutions is repeated until a termination condition is met. 

\subsection{Selection Policies}
\label{sec:SP}
This is a process of selecting individuals from population as parents to recombine and create off-springs for the next generation. It is very crucial for the convergence rate of GA and fittest parents drive to better solutions. For a high success of GA maintaining good diversity over population is extremely crucial, and one undesirable condition is premature convergence. Regarding selection methods to select learning paths from the population, then apply recombination operators to them, for the experimental phase we have developed these policies:

\begin{enumerate}
\item \textbf{Tournament Selection} - this method selects the fittest candidates from the current generation and passes them to the next generation. The selection of some potential candidates happens randomly, then the candidate with the highest fitness is selected from them. The steps we have followed to build this selection method are:
\begin{enumerate}
    \item Select k elements from the population.
    \item Select the best individual from the k elements in step 1.
\end{enumerate}

Implementation of Tournament Selection into our algorithm is like this:
\begin{lstlisting}[label={lst:tournament-selection}, caption = {Tournament Selection}, captionpos=b, language = Java , frame = single, escapeinside={(*@}{@*)}]
public LearningPath tournamentSelection(List<LearningPath> pop){
    List<LearningPath> selected = pickNRandomElements(pop, this.tS);
    return (LearningPath) Collections.max(selected);
}

\end{lstlisting}

The best individual selected becomes parent, on which the recombination operators will be applied. In case we need to have two parents to apply the recombination, as in the case of crossover, then the process is repeated.

\item \textbf{Roulette Selection} - the other method we used to select individuals from population, is Roulette Selection. The process followed to apply this selection approach is described in chapter 4, while our implementation is presented in Code Snippet 5. This function calculates the sum of all fitness values, then for each candidate in the population determines the probability of being selected. In the end the partial sum of individuals in population is calculated and the individual for which partial sum exceeds total sum is chosen.

\begin{lstlisting}[label={lst:roulette-selection}, caption = {Roulette Selection}, captionpos=b, language = Java , frame = single, escapeinside={(*@}{@*)}]
public LearningPath rouletteSelection(List<LearningPath> pop){
    double tS = 0.0;
    LearningPath lp = null;
    for (int i = 0; i < pop.size(); i++) {
        totalSum += pop.get(i).getFitness();
    }

    for (int i = 0; i < pop.size(); i++) {
        pop.get(i).setLearningPathProb(tS, pop.get(i).getFitness());
    }

    double partialSum = 0;
    double roulette = (double) (Math.random());

    for (int i = 0; i < pop.size(); i++) {
        partialSum += pop.get(i).learningPathProbability;
        if (partialSum >= roulette) {
            lp = pop.get(i);
            break;
        }
    }
    return lp;
}


\end{lstlisting}

\end{enumerate}

\subsection{Recombination Operators}
\label{sec:RC}
Since GAs apply techniques for combining the chromosomes of different individuals, with the sole purpose of creating young individuals that are likely to have higher quality than current individuals, we have developed these techniques to use:
\begin{enumerate}
\item \textbf{SWAP Mutation} - This is a technique used for mutation, which makes chromosome changes in a single individual. This is a simple technique to change an individual's genetic material and is applied through these two simple steps: 
\begin{enumerate}
\item Select two positions of chromosomes,
\item Swap their values.
\end{enumerate}
The Figure \ref{fig:swap-mutation} shows the swap mutation application process, which consists of these two steps above.

\begin{figure}[h]
    \begin{center}
        \includegraphics[width=100mm]{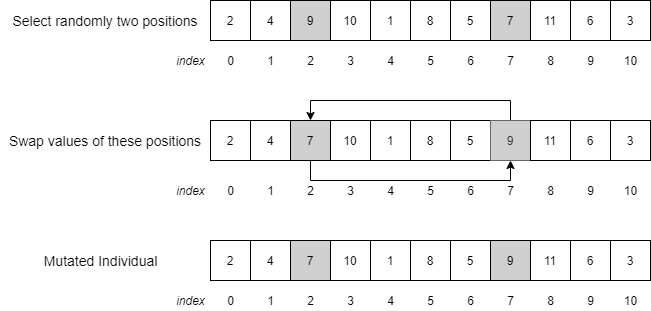}
        \caption{SWAP Mutation}
        \label{fig:swap-mutation}
    \end{center}
\end{figure}

\item \textbf{PMX} - Partially-Mapped Crossover for some problems offers better performance than most other crossover techniques. It builds an offspring by choosing a subsequence from one parent and preserving the order and position of as many alleles as possible from the other parent. The subsequence is selected by choosing two random cut points, which serve as boundaries for the swapping operations. The steps we have followed to perform this operation are:
\begin{enumerate}
\item Randomly select a subsequence from parent 1 and copy them direct to the child. Keep note the indexes of this segment.
\item Select each value which is not already been copy to the child looking in the same segment positions in parent 2.
\begin{enumerate}
    \item For each of these values: 
    \begin{enumerate}
        \item Note the index of this value in parent 2. From parent 1 in the same position locate this value.
        \item Locate the same value from parent 2.
        \item If the index of this value in parent 2 is part of the subsequence, using this value go to step \textbf{A}.
        \item Else insert step \textbf{i}’s value into the child to this position.
    \end{enumerate}
\end{enumerate}
\item Copy the remaining positions from parent 2 to the child if there is any.
\end{enumerate}
We use this function by sending two solutions from the population, which are selected randomly and are known as parents, and then in these parents the process described above is applied, thus creating two new solutions, which are placed in the population and are known as children. Figure \ref{fig:pmx-crossover} visually presents this process and visually describes the steps we have followed to achieve the generation of child solutions.

\begin{figure}[h]
    \begin{center}
        \includegraphics[width=90mm]{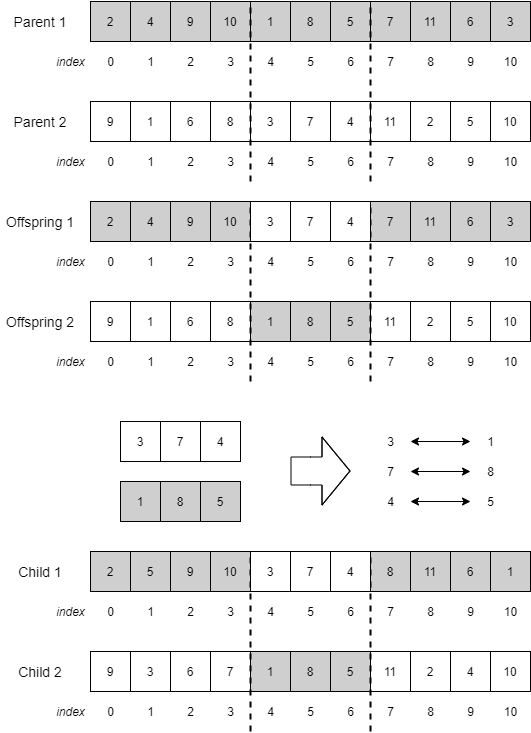}
        \caption{PMX Crossover}
        \label{fig:pmx-crossover}
    \end{center}
\end{figure}

\item \textbf{CYCLE} - Cycle Crossover is the other genetic operator we have used in our approach. It attempts to create children from the parents where every position is occupied by a corresponding chromosome from one of the parents. As the name itself indicates, this operator is based on closed loops formed by elements from two individuals selected by the population. To implement cycle crossover into our proposed algorithm, we have followed these steps:
\begin{enumerate}
    \item Create a cycle of chromosomes from first parent, in the following way:
    \begin{enumerate}
        \item Start with the first chromosome of first parent.
        \item Check at the chromosome with at the same position in second parent.
        \item Go to the position with the same chromosome in the first parent.
        \item Add this chromosome to the cycle.
        \item Repeat steps \textbf{ii} to \textbf{iv} until arrival at the first chromosome of first parent.
    \end{enumerate}
    \item Put the chromosomes of the cycle in the first child on the positions they have in the first parent.
    \item Take next cycle from the second parent. 
\end{enumerate}

These steps are also presented visually in Figure \ref{fig:cycle-crossover}.

\begin{figure}[h]
    \begin{center}
        \includegraphics[width=160mm]{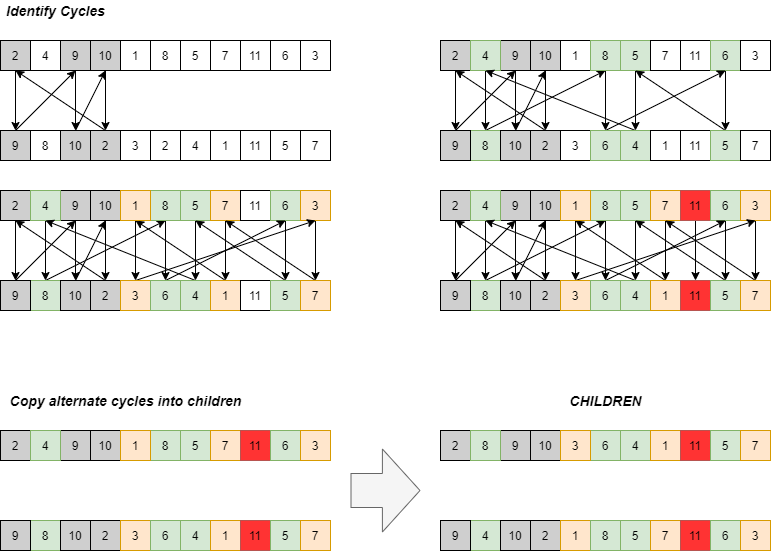}
        \caption{CYCLE Crossover}
        \label{fig:cycle-crossover}
    \end{center}
\end{figure}

The process of finding the cycles is described in the upper part of the figure and at the end each formed cycle is filled with a color to distinguish it from the other cycles. Meanwhile, the bottom part illustrates the creation of young individuals (children) by placing the parts of the cycle in the positions they have in the first parent, while the next cycle is copied from the second parent.
\end{enumerate}

\subsection{Population Initialization}
\label{sec:PI}
There are several approaches and heuristic to initialize the population used by GA to start the process of finding optimal solutions from it. The simplest and most used approach in GA applications for population initialization is to randomly generate individuals and then put them in the population until a limited number of defined population size is reached. This is a dumb idea, but for most problems it works very well. Another way to initialize the population, is starting the process with solutions that have been previously evaluated and have a higher affinity than randomly generated solutions. For the experimental phase we have deployed three ways for population initialization, randomly, Hill Climbing and Simulated Annealing. These approaches are compared in the end, to test which one is providing better starting input for personalized learning paths. The code to randomly initialize population is given in Listing \ref{lst:randomly-initialization}. The first method is used to instantiate learning paths according to the population size, whereas the second method is used to create a list of integers, which represents the path of concepts to follow.

\begin{lstlisting}[label={lst:randomly-initialization}, caption = {Randomly Population Initialization}, captionpos=b, language = Java , frame = single, escapeinside={(*@}{@*)}]
public List<LearningPath> initialPopulation(){
    List<LearningPath> population = new ArrayList();
    for (int i = 0; i < this.populationSize; ++i){
        population.add(new LearningPath());
    }
    return population;
}

private List<Integer> randomLearningPath(){
    List<Integer> result = new ArrayList<Integer>();
    for (int i = 0; i < numberOfConcepts; i++){
        result.add(i);
    }
    Collections.shuffle(result);
    return result;
}

\end{lstlisting}

\begin{enumerate}
    \item \textbf{Hill Climbing} - is a local search algorithm which continuously moves towards of increasing fitness value to find the peak of the mountain that also is the best solution for the problem. The condition to terminate is when it reaches a peak where no neighbor has a higher value. The process we followed to build this algorithm and use it into our approach is:
    
    \begin{enumerate}
        \item Generate a randomly solution.
        \item Loop until the number of predefined iterations is reached.
        \item For each iteration get the neighbor of the randomly initialized solution.
        \begin{enumerate}
            \item If the neighbor fitness value is better than the randomly solution, then replace it with the neighbor solution.
            \item Else go to the step \textbf{b}.
        \end{enumerate}
        
        \item Add the remaining solution from above steps to the population.
        \item Repeat all the steps above until we reach the population size.
    \end{enumerate}
    
    \item \textbf{Simulated Annealing} - is an algorithm used for optimizing parameters in a model by imitating the Physical Annealing process. It is almost the same as Hill Climbing, but instead of picking the best move, it picks a random one and if the selected move improves the solution, then it is accepted, if not the algorithm makes the move anyway with a probability less than 1. This probability decreases exponentially as bad as the movement is. The mathematical model to calculate the acceptance probability is:
    
    \[ \textbf{\textit{P}} =  exp(\frac{e - e_{n}}{T}) \]
    
    Where \textbf{\textit{P}} is the acceptance probability, \textbf{e} is the current energy, \textbf{e}\textsubscript{n} is the new energy, whereas \textbf{T} is the parameter for controlling the annealing process. Implementation of this function is given in Listing \ref{lst:acceptance-probability}:
\begin{lstlisting}[label={lst:acceptance-probability}, caption = {Acceptance Probability Function}, captionpos=b, language = Java , frame = single, escapeinside={(*@}{@*)}]
public static double accProbability(double e, double nE, double T){
    if (e < nE) {
        return 1.0;
    }
    return Math.exp((e - nE) / T);
}

\end{lstlisting}
\end{enumerate}

\subsection{Data Collection and Preparation}
\label{sec:dcp}
\begin{figure}[h]
    \begin{center}
        \includegraphics[width=150mm]{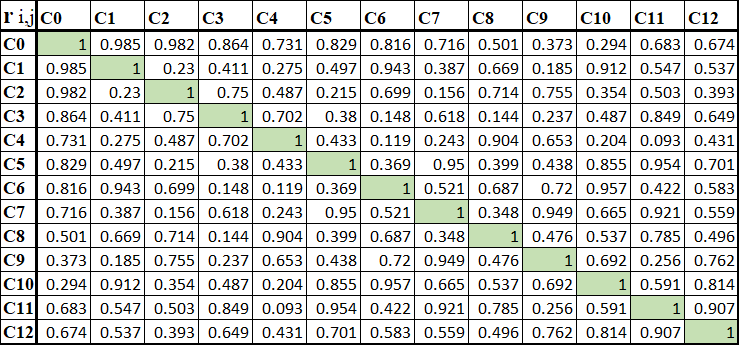}
        \caption{Learning Object Relation Degrees}
        \label{fig:relation-degrees}
    \end{center}
\end{figure}

The dataset used in the proposed model is adapted from \cite{pireva2017recommender}, which consists of learning objects within the Java course. From these data we extract a number of detailed information with respect to each learning object, such as: title, description, next learning object, relation degree with the next learning object, duration or granularity, difficulty value, rating by individual learner, overall rating, to name a few. However, the information that the objective function of the algorithm will accept as input for evaluating a solution are the following elements: granularity, individual rating, difficulty, and relation degree of learning objects. 
Since the algorithm needs the relation degree values for each learning object with the others, then these values are prepared in an excel sheet (see Figure \ref{fig:relation-degrees}) , from where the algorithm initially reads and stores them into a two-dimensional array.  

\section{Results and Discussion}
\label{sec:res}
This section provides the simulation results and observations obtained while performing the experiment of using our proposed GA based model on the data presented in section \ref{sec:mde}. The course unit “Control Flow and Conditionals” from “Introduction to Java Programming” course, is used to generate personalized learning paths, including different levels of difficulty, granularity, rating, and relation degrees between course concepts. The learning path for this course unit, that a student would follow in the traditional way of lecturing would be listed as in the Table \ref{lps-table}, and its fitness value is 41.68.

\setlength{\tabcolsep}{13.5pt}
\renewcommand{\arraystretch}{1.5}
\setlength{\arrayrulewidth}{0.2mm}
\begin{table}[h]
\begin{tabular}{|l|l|l|l|l|l|}
\hline
\textbf{ID} & \textbf{Title}               & \textbf{Difficulty} & \textbf{Granularity} & \textbf{Rating} & \textbf{Relation Degree} \\ \hline
0           & Introducing   Control Flow   & 1                   & 5                    & 6               & 0.985                    \\ \hline
1           & Decision   Making            & 1.65                & 10                   & 7               & 0.982                    \\ \hline
2           & If   Statement               & 2.15                & 12.5                 & 6.5             & 0.864                    \\ \hline
3           & Variable   Scope             & 2.35                & 15                   & 5.5             & 0.731                    \\ \hline
4           & Else   Statement             & 2.95                & 15                   & 7               & 0.829                    \\ \hline
5           & Else   If                    & 3.15                & 13                   & 8               & 0.816                    \\ \hline
6           & Multiple   Else Ifs          & 3.95                & 18                   & 6.5             & 0.716                    \\ \hline
7           & Boolean   Expressions        & 2.75                & 8                    & 7.5             & 0.501                    \\ \hline
8           & Logical   Operators          & 2.15                & 10                   & 7               & 0.373                    \\ \hline
9           & Logical   Operators Practice & 2.25                & 7.5                  & 8               & 0.294                    \\ \hline
10          & Nested   If Statements       & 4.55                & 20                   & 8.5             & 0.683                    \\ \hline
11          & Switch   Statement           & 4.15                & 20                   & 8.5             & 0.674                    \\ \hline
12          & Conclusion                   & 1.85                & 14                   & 9               & -                        \\ \hline
\end{tabular}
\caption{Learning Path Sequence}
\label{lps-table}
\end{table}

Assume that the student before starting to attend this unit performs a pretest and answers incorrectly in all learning objectives, and Figure \ref{fig:relation-degrees} illustrates the concept relation degrees of corresponding concept that student gives incorrect answers, then based on the relation degrees table and additional data related to a learning object extracted from dataset, the modeled algorithm can construct high quality personalized paths according to the designed objective function. Improving the quality of learning paths occurs throughout the evolution of generations, but to observe the trend of quality improvement we will compare all possible combinations of recombination operators, selection techniques and initialization approaches, that are developed as part of this experiment. The results of each combination used from the algorithm is presented in the Table \ref{rpc-table} , associated with best performing learning sequence, first fitness value found in the first generation and best fitness value after reaching 150 generations as defined in algorithm configuration parameters.

\setlength{\tabcolsep}{6.5pt}
\renewcommand{\arraystretch}{1.5}
\setlength{\arrayrulewidth}{0.2mm}
\begin{table}[h]
\begin{tabular}{|l|l|l|l|l|l|}
\hline
\multicolumn{1}{|c|}{\textbf{\begin{tabular}[c]{@{}c@{}}Selection   \\ Policy\end{tabular}}} & \multicolumn{1}{c|}{\textbf{\begin{tabular}[c]{@{}c@{}}Population   \\ Initialization \\ Approach\end{tabular}}} & \multicolumn{1}{c|}{\textbf{\begin{tabular}[c]{@{}c@{}}Crossover   \\ Type\end{tabular}}} & \multicolumn{1}{c|}{\textbf{\begin{tabular}[c]{@{}c@{}} | First \\ Fitness | \end{tabular}}} & \multicolumn{1}{c|}{\textbf{\begin{tabular}[c]{@{}c@{}} | Last \\ Fitness | \end{tabular}}} & \multicolumn{1}{c|}{\textbf{\begin{tabular}[c]{@{}c@{}}Learning   \\ Path \\ Sequence\end{tabular}}} \\ \hline
Tournament                                                                                   & Random                                                                                                           & PMX                                                                                       & 41.59                                                                                    & 45.14                                                                                   & {[}0,   2, 9, 7, 12, 1, 6, 10, 5, 11, 3, 4, 8{]}                                                     \\ \hline
Tournament                                                                                   & Random                                                                                                           & CYCLE                                                                                     & 41.63                                                                                    & 45.47                                                                                   & {[}0,   2, 3, 4, 8, 1, 6, 10, 12, 11, 5, 7, 9{]}                                                     \\ \hline
Tournament                                                                                   & Hill   Climbing                                                                                                  & PMX                                                                                       & 42.13                                                                                    & 44.73                                                                                   & {[}0,   2, 8, 6, 12, 11, 3, 4, 9, 7, 5, 10, 1{]}                                                     \\ \hline
Tournament                                                                                   & Hill   Climbing                                                                                                  & CYCLE                                                                                     & 42.17                                                                                    & 45.34                                                                                   & {[}0,   1, 6, 10, 12, 5, 7, 11, 3, 2, 8, 4, 9{]}                                                     \\ \hline
Tournament                                                                                   & Simulated   Annealing                                                                                            & PMX                                                                                       & 43.44                                                                                    & 44.9                                                                                    & {[}0,   3, 2, 8, 4, 1, 6, 9, 7, 11, 5, 10, 12{]}                                                     \\ \hline
Tournament                                                                                   & Simulated   Annealing                                                                                            & CYCLE                                                                                     & 43.89                                                                                    & 44.71                                                                                   & {[}0,   4, 8, 2, 6, 1, 10, 5, 11, 7, 9, 12, 3{]}                                                     \\ \hline
Roulette                                                                                     & Random                                                                                                           & PMX                                                                                       & 38.04                                                                                    & 44.48                                                                                   & {[}1,   2, 3, 0, 9, 11, 5, 8, 4, 7, 12, 10, 6{]}                                                     \\ \hline
Roulette                                                                                     & Random                                                                                                           & CYCLE                                                                                     & 38.4                                                                                     & 43.23                                                                                   & {[}8,   7, 1, 2, 3, 5, 4, 10, 11, 12, 0, 9, 6{]}                                                     \\ \hline
Roulette                                                                                     & Hill   Climbing                                                                                                  & PMX                                                                                       & 38.98                                                                                    & 43.72                                                                                   & {[}7,   0, 1, 8, 3, 12, 5, 9, 6, 11, 10, 2, 4{]}                                                     \\ \hline
Roulette                                                                                     & Hill   Climbing                                                                                                  & CYCLE                                                                                     & 40.66                                                                                    & 44.99                                                                                   & {[}0,   2, 9, 7, 12, 8, 4, 3, 11, 5, 10, 6, 1{]}                                                     \\ \hline
Roulette                                                                                     & Simulated   Annealing                                                                                            & PMX                                                                                       & 40.47                                                                                    & 43.9                                                                                    & {[}3,   9, 10, 1, 0, 12, 5, 6, 7, 11, 8, 2, 4{]}                                                     \\ \hline
Roulette                                                                                     & Simulated   Annealing                                                                                            & CYCLE                                                                                     & 38.83                                                                                    & 44.29                                                                                   & {[}4,   1, 10, 5, 9, 11, 8, 2, 6, 7, 0, 3, 12{]}                                                     \\ \hline
\end{tabular}

\caption{Results of Possible Combinations}
\label{rpc-table}
\end{table}

From the results presented in Table \ref{rpc-table}, the combination that has resulted in the learning path with the poorest quality (43.23) is roulette selection with random initialization and cycle crossover, while the combination that has resulted in the solution with the highest quality (45.47) is tournament selection with random initialization and cycle crossover. What can we claim from these results compared to the quality of the traditional learning path, is that even the quality of the weakest learning path generated by our GA approach is better than the quality of the traditional learning path, with a difference of 3.59\%, while the best solution generated compared to the traditional learning path is better for 8.34\%. In the following sections are present the results obtained by comparing different techniques of population initialization, selection policies and crossover methods that is used in our approach. The following section presents the results of the comparison between randomly population initialization and using other heuristics, such as Hill Climbing and Simulated Annealing.

\subsection{Results based on Population Initialization Approach}
\label{sec:rpia}

\begin{figure}[h]
    \begin{center}
        \includegraphics[width=120mm]{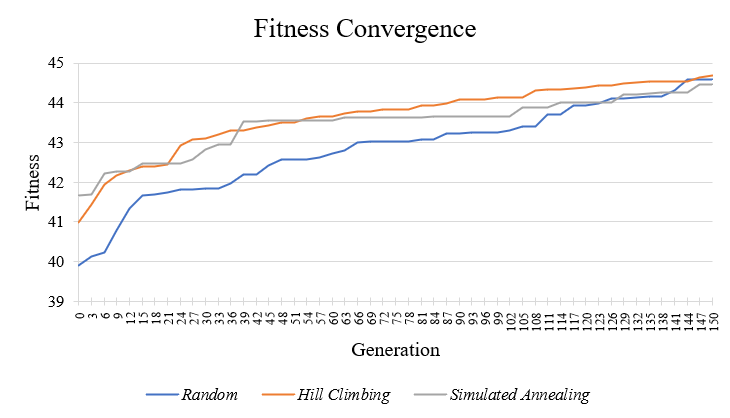}
        \caption{Path's Fitness Convergence by Population Initialization Approach}
        \label{fig:FC-Population}
    \end{center}
\end{figure}

In this section are discussed the results and evolution of the learning path using as a starting generation a population created by random solutions, as opposed to the use of a population generated through known heuristics, such as Hill Climbing and Simulated Annealing. As we can conclude from Figure \ref{fig:FC-Population}, in the first generation created, SA provided the most qualitative solution, then continued with HC and followed by filling the generation with random solutions. In the first iterations of the algorithm this trend continued, then changing the course of evolution of solutions from the generation created after iteration of 51 using the HC technique. Upon reaching the final iteration, which was also the termination condition, the quality of the best solution in the last generation was almost the same in all the techniques used. From these results we can conclude that the use of population initialization techniques affects only the first generation, because only at that moment the approach of population initialization is applied, then the evolution of the solution quality depends on other factors, such as selection policies and recombination operators

\subsection{Results based on Selection Policy}
\label{sec:rsp}
\begin{figure}[h]
    \begin{center}
        \includegraphics[width=110mm]{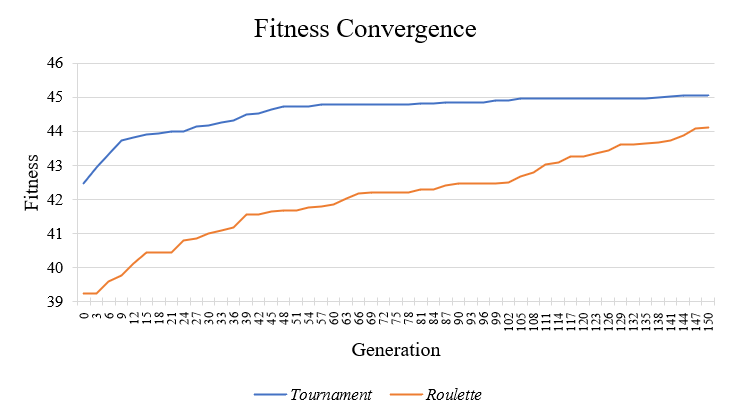}
        \caption{Path's Fitness Convergence by Selection Policies}
        \label{fig:FC-Selection}
    \end{center}
\end{figure}

To see the impact of the selection policies of individuals from the current generation and then their use to create the new generation, there is analyzed the performance of the two policies developed in our approach. From the graph depicted in Figure \ref{fig:FC-Selection} there is visible that at the initial development stage, Tournament Selection performed much better than Roulette Selection. In the first generation the best solution from using TS has had a fitness value of 42.74, while RS produced the best solution with fitness value of 39.23. An interesting pattern that can be noticed is that TS after the iteration of 60, remains in the plateau, making the evolution of fitness value for the best solutions in the next generations to be minor. While the other selection technique after each iteration tends to improve the evolution of the solution by increasing the value of fitness after each new generation until the last iteration. Although this upward trend of RS seems promising, in the end it still fails to produce better solutions compared to TS. In the last iteration, the solution found using TS has a fitness value of 45.04, while the use of RS has provided a solution with fitness of 44.10.

\subsection{Results based on Crossover Type}
\label{sec:rct}
\begin{figure}[h]
    \begin{center}
        \includegraphics[width=120mm]{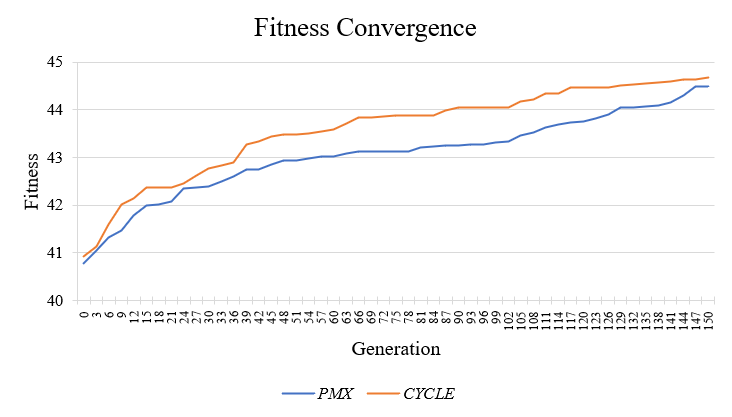}
        \caption{Path's Fitness Convergence by Crossover Methods}
        \label{fig:FC-Crossover}
    \end{center}
\end{figure}

From the results of comparing the performance of the implemented crossover types, the first generations, both types produced solutions with approximate fitness value. With the increase in the number of iterations and the creation of new generations, cycle crossover has shown higher potential in chromosome combination and higher degree of search space exploitation, thus providing more potential solutions for use in creating new generations. However, as we can conclude from the visualization of the results in Figure \ref{fig:FC-Crossover}, although cycle crossover has shown better performance for a small nuance throughout generation life-cycle, still in the last iteration the use of both techniques has provided valuable learning pathways, and their fitness value is almost the same.

\section{Conclusion and future work}
\label{sec:conclusion}

In this research paper is introduced a new genetic algorithm hybrid approach that is able to find and generate personalized learning trajectories considering course difficulty level, duration, rating, and relation degrees between learning objects. From the results of the experimental phase, the authors conclude that the proposed algorithm is able to find better solutions compared to the traditional path, which is uniformly modeled for all students (presented in Results section). Furthermore, the authors conclude that the most optimal solution was generated by using tournament selection with random population initialization and cycle crossover, while the combination with the weakest score was roulette selection with random initialization and cycle crossover. Also, from the results achieved, we can observe that the objective function presented in this paper corresponds to the problem of composing personalized learning paths, as it shows convergence with increasing number of iterations. From the individual comparison of population initialization approaches, we can say that Simulated Annealing, as one of the most advanced techniques for addressing the problem of population initialization has contributed on the creation of first generations with solutions closer to the optimum. While from the selection policies with the best performance has resulted Tournament Selection, although after a certain number of iterations somehow stalled on the plateau, then continuing with the provision of solutions with small improvements in quality. Regarding the comparison of crossover methods, between PMX and cycle, there was no very distinct difference in their performance, resulting in the crossover cycle being dominant for a very small nuance.
This paper makes three crucial contributions:
\begin{enumerate}
    \item Review of the preliminary literature on the use of genetic algorithm in the composition of personalized learning paths based on the learner’s profile and the attributes of the course topics.
    \item Modeling and developing a hybrid approach based on genetic algorithm of the course topics by encountering the duration, rating, and the relation degree between topics.
    \item Comparison of performance from the development of different techniques for population initialization, selection and recombination.
\end{enumerate}

With respect to future work,this approach can be extended including additional attributes related to the learner profile, such as prior knowledge background and modeling of an objective function that includes these additional attributes. Another possibility of improvement is the provision of new techniques for population initialization, selection of individuals to create new generations and various methods for combining genetic material, as well as tuning several configuration parameters, such as increasing the number of generations, population size, and tuning the probability for recombination operators.

\bibliographystyle{unsrt}  
\bibliography{references}  

\begin{thebibliography}{10}

\bibitem{pireva2017representation}
Krenare Pireva, Petros Kefalas, and Ioanna Stamatopoulou.
\newblock Representation of learning objects in cloud e-learning.
\newblock In {\em 2017 8th International Conference on Information,
  Intelligence, Systems \& Applications (IISA)}, pages 1--6. IEEE, 2017.

\bibitem{jih1996impact}
Hueyching~Janice Jih.
\newblock The impact of learners' pathways on learning performance in
  multimedia computer aided learning.
\newblock {\em Journal of network and computer applications}, 19(4):367--380,
  1996.

\bibitem{thomas2016future}
Susan Thomas.
\newblock Future ready learning: Reimagining the role of technology in
  education. 2016 national education technology plan.
\newblock {\em Office of Educational Technology, US Department of Education},
  2016.

\bibitem{karampiperis2005adaptive}
Pythagoras Karampiperis and Demetrios Sampson.
\newblock Adaptive learning resources sequencing in educational hypermedia
  systems.
\newblock {\em Journal of Educational Technology \& Society}, 8(4):128--147,
  2005.

\bibitem{pireva2014cloud}
Krenare Pireva, Petros Kefalas, Dimitris Dranidis, Thanos Hatziapostolou, and
  Anthony Cowling.
\newblock Cloud e-learning: A new challenge for multi-agent systems.
\newblock In {\em Agent and Multi-Agent Systems: Technologies and
  Applications}, pages 277--287. Springer, 2014.

\bibitem{pireva2017recommender}
Krenare Pireva and Petros Kefalas.
\newblock A recommender system based on hierarchical clustering for cloud
  e-learning.
\newblock In {\em International Symposium on Intelligent and Distributed
  Computing}, pages 235--245. Springer, 2017.

\bibitem{gascuena2006domain}
Jos{\'e}~Manuel Gascue{\~n}a, Antonio Fern{\'a}ndez-Caballero, and Pascual
  Gonz{\'a}lez.
\newblock Domain ontology for personalized e-learning in educational systems.
\newblock In {\em Sixth IEEE International Conference on Advanced Learning
  Technologies (ICALT'06)}, pages 456--458. IEEE, 2006.

\bibitem{de2007competency}
Luis de~Marcos, Carmen Pag{\'e}s, Jose-Javier Martinez, and Jose-Antonio
  Gutierrez.
\newblock Competency-based learning object sequencing using particle swarms.
\newblock In {\em 19th IEEE International Conference on Tools with Artificial
  Intelligence (ICTAI 2007)}, volume~2, pages 111--116. IEEE, 2007.

\bibitem{kwasnicka2008learning}
Halina Kwasnicka, Dorota Szul, Urszula Markowska-Kaczmar, and Pawel~B
  Myszkowski.
\newblock Learning assistant-personalizing learning paths in e-learning
  environments.
\newblock In {\em 2008 7th Computer Information Systems and Industrial
  Management Applications}, pages 308--314. IEEE, 2008.

\bibitem{kumar2005rating}
Vive Kumar, John Nesbit, and Kate Han.
\newblock Rating learning object quality with distributed bayesian belief
  networks: the why and the how.
\newblock In {\em Fifth IEEE International Conference on Advanced Learning
  Technologies (ICALT'05)}, pages 685--687. IEEE, 2005.

\bibitem{neil2005using}
Martin Neil, Norman Fenton, and Manesh Tailor.
\newblock Using bayesian networks to model expected and unexpected operational
  losses.
\newblock {\em Risk Analysis: An International Journal}, 25(4):963--972, 2005.

\bibitem{heckerman2008tutorial}
David Heckerman.
\newblock A tutorial on learning with bayesian networks.
\newblock {\em Innovations in Bayesian networks}, pages 33--82, 2008.

\bibitem{eiben2003introduction}
Agoston~E Eiben, James~E Smith, et~al.
\newblock {\em Introduction to evolutionary computing}, volume~53.
\newblock Springer, 2003.

\bibitem{jebari2011genetic}
Khalid Jebari, Abdelaziz El~Moujahid, Abdelaziz Bouroumi, and Aziz Ettouhami.
\newblock Genetic algorithms for online remedial education based on competency
  approach.
\newblock In {\em 2011 International Conference on Multimedia Computing and
  Systems}, pages 1--6. IEEE, 2011.

\bibitem{huang2007constructing}
Mu-Jung Huang, Hwa-Shan Huang, and Mu-Yen Chen.
\newblock Constructing a personalized e-learning system based on genetic
  algorithm and case-based reasoning approach.
\newblock {\em Expert Systems with applications}, 33(3):551--564, 2007.

\bibitem{chen2008intelligent}
Chih-Ming Chen.
\newblock Intelligent web-based learning system with personalized learning path
  guidance.
\newblock {\em Computers \& Education}, 51(2):787--814, 2008.

\bibitem{bhaskar2010genetic}
Manju Bhaskar, Minu~M Das, T~Chithralekha, and S~Sivasatya.
\newblock Genetic algorithm based adaptive learning scheme generation for
  context aware e-learning.
\newblock {\em International Journal on Computer Science and Engineering},
  2(4):1271--1279, 2010.

\bibitem{clement2000model}
John Clement.
\newblock Model based learning as a key research area for science education.
\newblock {\em International Journal of Science Education}, 22(9):1041--1053,
  2000.

\bibitem{de2011genetic}
Luis de~Marcos, Jose-Javier Martinez, Jose-Antonio Gutierrez, Roberto Barchino,
  Jose-Ramon Hilera, Salvador Oton, and Jos{\'e}-Mar{\i}a Guti{\'e}rrez.
\newblock Genetic algorithms for courseware engineering.
\newblock {\em International Journal of Innovative Computing, Information and
  Control}, 7(7):1--27, 2011.

\bibitem{hong2005personalized}
Chin-Ming Hong, Chih-Ming Chen, and Mei-Hui Chang.
\newblock Personalized learning path generation approach for web-based
  learning.
\newblock In {\em 4th WSEAS Int. Conf. On E-ACTIVITIES, Miami, Florida, USA},
  pages 62--68, 2005.

\bibitem{zaporozhko2018genetic}
Veronika~V Zaporozhko, Irina~P Bolodurina, and Denis~I Parfenov.
\newblock A genetic-algorithm approach for forming individual educational
  trajectories for listeners of online courses.
\newblock In {\em In Proceedings of Russian Federation \& Europe
  Multidiciplinary Symposium on Computer Science and ICT}, 2018.

\bibitem{bellafkih2010adaptive}
Mostafa Bellafkih.
\newblock Adaptive e-learning using genetic algorithms.
\newblock In {\em IJCSNS International Journal of Computer Science and Network
  Security}, volume~10, 2010.

\bibitem{holland1992adaptation}
John~Henry Holland et~al.
\newblock {\em Adaptation in natural and artificial systems: an introductory
  analysis with applications to biology, control, and artificial intelligence}.
\newblock MIT press, 1992.

\bibitem{rothlauf2009representations}
Franz Rothlauf.
\newblock Representations for evolutionary algorithms.
\newblock In {\em Proceedings of the 11th Annual Conference Companion on
  Genetic and Evolutionary Computation Conference: Late Breaking Papers}, pages
  3131--3156, 2009.

\end{thebibliography}



\end{document}